\documentclass[10pt,twocolumn,letterpaper]{article}

\usepackage[article]{cvpr}

\usepackage{graphicx}
\usepackage{amsmath}
\usepackage{amssymb}
\usepackage{booktabs}
\usepackage{ctable}
\usepackage{multirow}
\usepackage{array}
\usepackage{colortbl}
\usepackage{makecell}

\definecolor{cvprblue}{rgb}{0.21,0.49,0.74}
\usepackage[pagebackref,breaklinks,colorlinks,citecolor=cvprblue]{hyperref}

\graphicspath{{img/}{../img/}{./}{../}}

\def\paperID{}
\def\confName{CVPR}
\def\confYear{}

\begin{document}

\title{Towards Robust Training in NNGPT AutoML Pipeline: A Loss--Optimizer Pairing Selection Study }

\author{Anton Abramochkin{\thanks{Corresponding author: anton.abramochkin@stud-mail.uni-wuerzburg.de}}, \hspace{0.2cm} Radu Timofte, \hspace{0.2cm} Dmitry Ignatov\\
\small{Computer Vision Lab, CAIDAS \& IFI, University of W\"urzburg, Germany}}
\maketitle

\begin{abstract}
The choice of loss function and optimizer is an important decision, that
shapes further model training. Yet automated architecture search
pipelines (AutoML) benefits significantly more from the optimal pairing selection and vice versa.
This paper investigates whether a single recipe is sufficient for
heterogeneous architecture pools, or whether the optimal pairing varies
across structurally diverse models. We conduct a systematic empirical study of all $3 \times 6 = 18$
combinations of six optimizers (SGD+Momentum, Adam, AdamW, RMSprop, Adagrad, Adadelta), paired with three
loss functions: Cross-Entropy (CEL), Negative
Log-Likelihood (NLL), and the recently introduced genetically evolved NGL loss across the
base models presented in LEMUR heterogeneous architecture pool
on six image classification datasets (CelebA-Gender, CIFAR-10, CIFAR-100, ImageNette, MNIST, SVHN).
The 18 loss--optimizer configurations are applied to each of the 33 compatible base architectures taken from the LEMUR pool, resulting in 594 variants that were generated fully automatically by a source-level injection pipeline
and evaluated under fixed hyperparameters, ensuring that observed
accuracy differences are attributable solely to the loss--optimizer
pairing. 
Our results confirm that no single pairing is universally optimal.
Cross-Entropy with Adam or AdamW is the most robust choice across
architecture families and datasets.
NGL is a competitive alternative to CEL on standard convolutional
classifiers, but only when paired with adaptive optimizers; it
degrades substantially with SGD or accumulation-based methods.
Adagrad and Adadelta consistently underperform under fixed
hyperparameters regardless of loss function, highlighting their
sensitivity to learning rate tuning.
These findings provide actionable guidance for loss--optimizer
selection within NNGPT Framework.
\end{abstract}

\section{Introduction}
\label{sec:intro}

Whether that choice of a certain Loss--Optmizer pairing expands on structurally diverse architectures remains unexamined.

Within the NNGPT framework~\cite{ABrain.NNGPT,ABrain.HPGPT}, large language models continuously generate, train, 
and evaluate novel neural architectures~\cite{ABrain.NN-Captioning_2025,ABrain.Prompt,ABrain.NNGPT-Fractal,ABrain.Transform,ABrain.CV_Channel,ABrain.DeltaNAS,ABrain.NN-RAG} 
to identify high-performing solutions across diverse environments, including edge devices~\cite{ABrain.NN-Lite,ABrain.MobileAgeNet,ABrain.MobileDenoising}. 
During this process, hyperparameters~\cite{Aboudeshish2025augmentation}, network layers, and other structural components are extensively optimized; however, the loss function and optimizer typically remain fixed throughout the training pipeline.
Whether different architectures respond differently to the same pairing has not been systematically studied yet.

The problem is particularly acute for NNGPT Framework.
The LEMUR dataset~\cite{ABrain.NN-Dataset,ABrain.LEMUR2} contains a broad
range of architectures with distinct structural properties including standard
classifiers such as DenseNet, AirNext, and DarkNet; dual-path networks
(DPN68, DPN107, DPN131); segmentation backbones (FCN8s, LRASPP); mixture-of-experts
models (MoEv2); image captioning models with sequence decoders
(ConvNeXtTransformer, ResNetTransformer, RESNETLSTM); and complex-valued
networks (ComplexNet).
Understanding when and why a pairing succeeds or fails is a
practically important knowledge, and must be taken into account 
for further architectural explorations within the framework.

The emergence of NGL (Next Generation Loss)~\cite{Akhmedova2024NGL}, a loss function discovered via genetic programming,
raises a question about its robustness across different neural networks and optimizers. NGL has shown a competitive performance
that has matched or outperformed cross-entropy on CIFAR-10, CIFAR-100, Fashion-MNIST, and ImageNet-1k. However, no prior work has assessed 
whether NGL generalizes under fixed hyperparameters across structurally diverse models and multiple optimizers.
This study fills that gap, providing the first systematic evaluation of NGL within NNGPT. 

For each of the base neural network present within the LEMUR dataset~\cite{ABrain.NN-Dataset,ABrain.LEMUR2}
18 samples with injected loss function and optimizer were generated. 18 corresponds to the amount of pairings chosen for this study.

All variants in this study were trained and evaluated automatically via
the NNEval module of the NNGPT framework~\cite{ABrain.NNGPT}.
NNEval loads each architecture sample with modified loss function, runs a fixed training schedule on the target
dataset, and records top-1 accuracy as the evaluation metric, without
any per-architecture manual tuning.
This controlled evaluation pipeline ensures that observed accuracy
differences are attributable solely to the loss--optimizer pairing.

The contributions of this work are:
\begin{itemize}
  \item Empirical findings on the robustness of 18 loss--optimizer
    pairings across four classification benchmarks, providing
    actionable guidance for selecting training recipes in heterogeneous
    AutoML pipelines.
    \item More precise evaluation of NGL loss function. Received insights
    define clearer usecases in terms of architecture and optimizer pairing
    for the newly introduced loss function.
  \item A fully automated variant generation pipeline that injects any
    combination of loss and optimizer into existing LEMUR model source
    files, handling heterogeneous interfaces, spatial output models,
    and the NGL custom loss.
\end{itemize}

\section{Related Work}
\label{sec:Related}

This section reviews the components most relevant to our study:
the NNGPT framework and LEMUR dataset that form our experimental
infrastructure, the three loss functions selected for evaluation, 
including NGL, which capabilities are largely untested
and the six optimizers with which they are paired.

\subsection{NNGPT and LEMUR}

NNGPT~\cite{ABrain.NNGPT} is an LLM-based AutoML framework in which large
language models iteratively generate, train, and evaluate neural
architectures to discover high-performing designs.
The framework includes NNEval, an automated evaluation module that trains
each candidate architecture from scratch on a given dataset for a fixed
number of epochs and records the top-1 accuracy as the evaluation
signal~\cite{ABrain.HPGPT}.

The LEMUR dataset~\cite{ABrain.NN-Dataset,ABrain.LEMUR2} provides a
pool of neural architectures including standard image classifiers,
dual-path networks, segmentation models, captioning models,
mixture-of-experts, and lightweight edge models.
This architectural diversity of LEMUR provides an opportunity to study
how loss--optimizer pairings interact with heterogeneous model designs.
Evaluation results may be beneficial for further loss--optimizer pair
selection.

\subsection{NGL and Standard Loss Functions for Image Classification}

The choice of loss function shapes the gradient that is responsible for updating
the parameters. Loss function behaviour varies depending on both
model output and optimizer selection. 
We survey loss functions ranging from the well-established standards to recent proposals,
and select 3 functions that together represent this range for further evaluation.

NGL (Next Generation Loss) was discovered via genetic
programming~\cite{Akhmedova2024NGL}, a population-based evolutionary
algorithm that constructs loss expressions from elementary operators
and leaf nodes and iteratively recombines and mutates them.
The resulting function is:
\begin{equation}
  \mathcal{L}_{\mathrm{NGL}}(x, y)
  = \mathbb{E}\bigl[\exp(2.4092 - x - x{\cdot}\hat{y})
    - \cos(\cos(\sin(x)))\bigr],
\end{equation}
where $x = \mathrm{softmax}(\hat{f})$ and $\hat{y}$ is the one-hot
encoding of the ground-truth label.
Akhmedova and K\"{o}rber~\cite{Akhmedova2024NGL} evaluated NGL on
ResNet50 and InceptionV3 across seven datasets (Malaria, PCam,
Colorectal Histology, CIFAR-10, Fashion-MNIST, CIFAR-100,
Caltech~101), and further on ResNet101, ResNet152, Swin-T and Swin-S
on ImageNet-1k, reporting the same or better top-1 accuracy compared
to cross-entropy in all cases.
However, this evaluation covers only a small, manually selected set of
standard architectures and datasets; no prior work has assessed NGL across both more 
expanded list of neural networks and heterogeneous, automatically generated architecture 
pool or across multiple optimizer choices.

Cross-entropy loss (CEL) has proven itself as a reliable standard for image
classification tasks~\cite{Akhmedova2024NGL,MaoMohri2023CE}.
It combines log-softmax and negative log-likelihood into a single
numerically stable operation that directly maximises the posterior
probability of the correct class, providing smooth gradients
during training.

Negative Log-Likelihood Loss (NLLLoss) is the classical formulation
for training multi-class classifiers~\cite{pytorch}, where the
log-softmax activation is applied explicitly in the model's output
layer before the loss receives log-probabilities.
It is mathematically equivalent to CEL when used with log-softmax,
but is a distinct choice for architectures that already incorporate
an explicit log-softmax in their final layer.

Several other candidate loss functions were considered but excluded
due to fundamental incompatibilities with the LEMUR architecture pool, 
which encompasses exclusively image classification neural networks.
Multi-Margin Loss implements the multi-class SVM hinge
objective~\cite{CrammerSinger2001}, which is designed for margin-based
classifiers not present in the pool.
KL-Divergence Loss~\cite{Kullback1951KL} requires a full probability
distribution as the training target rather than a class index, which
standard single-label classification datasets do not provide~\cite{Goodfellow2016DL}.

\subsection{Optimization Algorithms}
We review the six optimizers evaluated in this work, covering
non-adaptive momentum-based methods, adaptive moment-based methods,
and adaptive methods with alternative accumulation strategies.

SGD with momentum, introduced by Polyak~\cite{Polyak1964}, accumulates
a velocity vector in the direction of persistent gradient flow, dampening
oscillations and accelerating convergence along low-curvature directions.
It remains a strong baseline in image classification: Choi et
al.~\cite{Choi2020optimizers} show that well-tuned SGD can match
adaptive methods, and it was the optimizer of choice for training
canonical vision models such as ResNet~\cite{He2016ResNet}.

Adam, proposed by Kingma and Ba~\cite{Kingma2015Adam}, maintains
exponential moving averages of both the gradient (first moment) and
the squared gradient (second moment) to compute per-parameter adaptive
learning rates, with bias-correction terms that stabilise updates
during early training.
AdamW, introduced by Loshchilov and Hutter~\cite{Loshchilov2019AdamW},
corrects a flaw in Adam's weight decay implementation: in Adam, L2
regularisation is applied to the gradient before the adaptive scaling,
which causes parameters with large gradient variance to be
underregularised; AdamW decouples weight decay from the gradient update
entirely, consistently improving generalisation over Adam.

RMSprop, introduced by Tieleman and Hinton~\cite{Tieleman2012RMSprop},
divides the learning rate by a running average of recent squared
gradient magnitudes, preventing the effective learning rate from
decaying to zero.
Adagrad, proposed by Duchi et al.~\cite{Duchi2011Adagrad}, accumulates
all past squared gradients to assign larger updates to infrequent
parameters and smaller updates to frequent ones, making it well-suited
for sparse features but prone to stagnation in deep networks
due to its monotonically shrinking step size.
Adadelta, introduced by Zeiler~\cite{Zeiler2012Adadelta}, addresses
this by replacing the full accumulation window with a decaying average
of recent squared gradients and recent squared parameter updates,
removing the need to set a global learning rate altogether.

Choi et al.~\cite{Choi2020optimizers} establish a formal inclusion hierarchy
among these families: SGD $\subset$ Momentum $\subset$ RMSprop and SGD $\subset$ Momentum $\subset$ Adam, meaning that with
sufficient hyperparameter tuning, more general optimizers
should never underperform their specialisations. They also
show that ``the metaparameter search space may be the single most important
factor explaining the rankings obtained by recent empirical
comparisons'', motivating our design
decision to hold all hyperparameters fixed and vary only the
loss--optimizer pair.

\section{Methodology}
\label{sec:Methodology}

Inspired by recent advancements in LLM-based neural architecture search
and evaluation~\cite{ABrain.NNGPT,ABrain.Architect,ABrain.HPGPT,ABrain.Feedback_Memory}
and leveraging the LEMUR pool of diverse
architectures~\cite{ABrain.NN-Dataset,ABrain.LEMUR2}, we designed a
controlled experimental study of loss--optimizer interactions across
heterogeneous neural architectures.
Rather than generating new architectures, we take an existing LEMUR
model as a fixed base and systematically substitute its loss function
and optimizer, producing a $3 \times 6$ grid of 18 training
configurations per architecture.
All variants are evaluated under identical conditions using the NNEval
module of the NNGPT framework~\cite{ABrain.NNGPT}, ensuring that
observed accuracy differences depend solely on the loss--optimizer pair.

\subsection{Experimental Setup}

All variants were trained under the unified default hyperparameter
configuration of NNEval~\cite{ABrain.NNGPT,ABrain.HPGPT}, listed in
Table~\ref{table:hparams}.
This fixed-hyperparameter design is motivated by the necessity of a
fairer permormance evaluation, which can be executed in the same point of hyperparameter space.
follows the recommendation of Choi et
al.~\cite{Choi2020optimizers}, who demonstrate that optimizer rankings
produced under unequal tuning budgets are not scientifically
meaningful.
Preserving all hyperparameters constant isolates the effect of the
loss--optimizer pairing from deviations introduced by hyperparameter
search, at the cost of not reporting the best achievable accuracy for
each individual pair.

\begin{table}[h]
  \centering
  \fontsize{8}{9.5}\selectfont
  \caption{Fixed default hyperparameters used in all training runs
    (NNEval defaults; \texttt{epoch} is overridden per experiment).}
  \label{table:hparams}
  \begin{tabular}{lc}
    \toprule
    \textbf{Hyperparameter} & \textbf{Value} \\
    \midrule
    Learning rate (LR) & 0.01 \\
    Batch size & 64 \\
    Dropout & 0.2 \\
    Momentum (SGD only) & 0.9 \\
    Input transform & \texttt{norm\_256\_flip} \\
    Stochastic depth prob. & 0.0 \\
    Attention dropout & 0.0 \\
    Norm epsilon ($\varepsilon$) & $10^{-5}$ \\
    Norm momentum & 0.1 \\
    Patch size & 0.125 \\
    \bottomrule
  \end{tabular}
\end{table}

\subsection{Variant Generation Pipeline}

Because the LEMUR architectures were contributed by independent authors,
injecting a new loss function or
optimizer cannot be done through a unified API call.
We therefore developed an automated source-level transformation pipeline
that treats each model file as a structured text artifact and rewrites
the relevant training assignments in place, without modifying the
architecture logic.

For each of the 18 loss--optimizer combinations, the pipeline locates
the loss and optimizer assignments inside the model's training setup,
replaces them with the target components, and propagates any
interface-level consequences. Where necessary, it also handles side effects
such as registering
optimizer-specific hyperparameters with the evaluation framework or
injecting the NGL class definition, which is not part of the standard
PyTorch library.
The result is a syntactically valid model file that is functionally
identical to the original except for the substituted training recipe.

This approach decouples the experimental grid from the architectural
diversity of the pool: adding a new loss or optimizer to the study
requires no per-model manual work, and the same pipeline scales to
any future extension of the LEMUR pool.

\section{Experiments}
\label{sec:Experiments}

\subsection{Experimental Protocol}

Training of computer vision models is performed using the
AI~Linux docker image \texttt{abrainone/ai-linux}%
\footnote{\url{https://hub.docker.com/r/abrainone/ai-linux}}
on NVIDIA GeForce RTX~3090/4090 24\,GB GPUs of the CVL Kubernetes
cluster at the University of Wueurzburg.
Each variant is trained for 5 epochs with all hyperparameters fixed
at the NNEval defaults (Table~\ref{table:hparams}); the only variables
are the loss function and optimizer.
Evaluation is fully automated via
NNEval~\cite{ABrain.NNGPT,ABrain.HPGPT}: each trained variant is
scored by top-1 accuracy on the held-out test split of the respective
dataset.

The evaluation was conducted on the original LEMUR base architectures
with substituted loss--optimizer pairs only, without any structural
modifications to the networks themselves.
In total, the study produced 594 variants: 18 loss--optimizer
configurations for each of the 33 base architectures in the compatible
evaluation set.
The evaluation covers six datasets: MNIST, SVHN, CIFAR-10, CIFAR-100,
ImageNette, and CelebA-Gender.
\subsection{Architecture Compatibility}
\label{sec:compat}

Not all LEMUR architectures are compatible with standard single-label
image classification training.
Prior to evaluation, we identified three categories of fundamental
architecture--dataset mismatches that cannot be resolved without
modifying the base architectures themselves and excluded the affected
models from the study.
Captioning models (ConvNeXtTransformer, ResNetTransformer, RESNETLSTM)
expect sequence-labelled batches and are incompatible with scalar class
labels.
ComplexNet requires complex-valued input tensors, which no standard
classification dataset provides.
MoEv2 hardcodes its flattened feature dimension to
$128 \times 4 \times 4 = 2048$ for $32\times 32$ inputs, causing a
matrix multiplication failure on all larger images.
The remaining architectures in the LEMUR pool form the evaluation set
used throughout this study.

\subsection{Evaluation Metrics}

The primary evaluation metric is top-1 accuracy on the held-out test
split of each dataset, as recorded by NNEval~\cite{ABrain.NNGPT}.
Since the six datasets differ substantially in difficulty, number of
classes, and image resolution, raw accuracy values are not
directly comparable across datasets.
A pairing that achieves 95\% on MNIST and 45\% on CIFAR-100 does not
necessarily indicate stronger performance on the former; the baseline
difficulty of each task must be taken into account when interpreting
results.
We therefore report accuracy within each dataset separately rather
than averaging across them, preserving the per-task signal.

We additionally report the per-dataset training success rate,
defined as the fraction of variants that complete training without
error, to characterise the compatibility of each loss--optimizer pairing
with the evaluated architecture pool.
Success rate varies considerably across datasets because architectures
with fixed spatial assumptions (e.g.\ hardcoded feature dimensions)
fail only on datasets whose image resolution differs from the
assumption, while architectures compatible with all resolutions
succeed uniformly.

While analysing loss functions and optimizers in isolation reveals
their individual contributions, the interaction between the two is
equally important: a loss function may perform well with one optimizer
and poorly with another, and these interaction effects would be
invisible in marginal plots.
The joint loss--optimizer accuracy matrix and per-dataset success rates
are therefore presented together with the detailed results in
Section~\ref{sec:results}.

\section{Results and Discussion}
\label{sec:results}

\begin{figure*}[t]
  \centering
  \begin{minipage}[b]{0.48\textwidth}
    \centering
    \includegraphics[width=\linewidth]{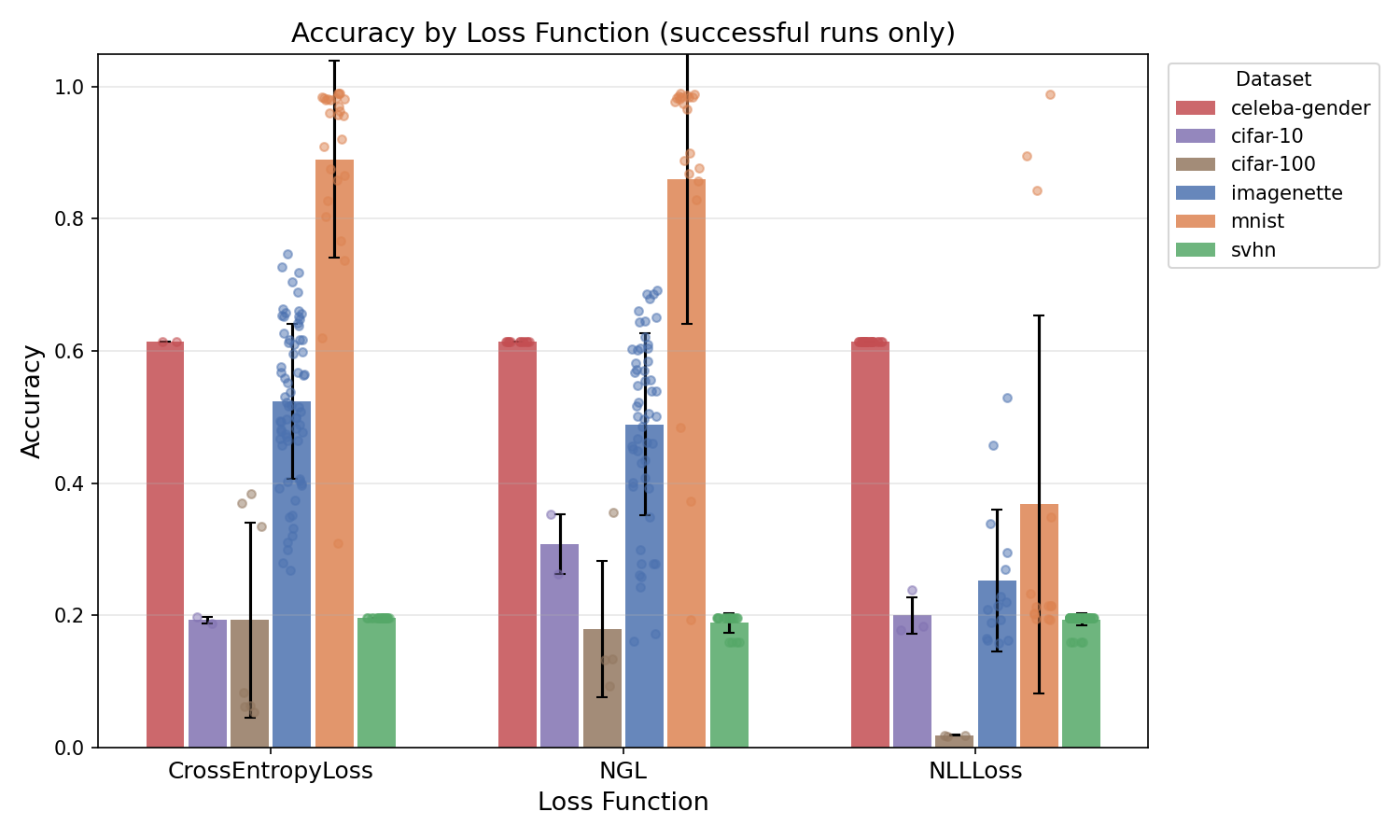}
    \caption{Top-1 accuracy by loss function. Each point
      represents one successfully trained variant. Box plots show
      average and interquartile range per loss function.}
    \label{fig:loss_acc}
  \end{minipage}\hfill
  \begin{minipage}[b]{0.48\textwidth}
    \centering
    \includegraphics[width=\linewidth]{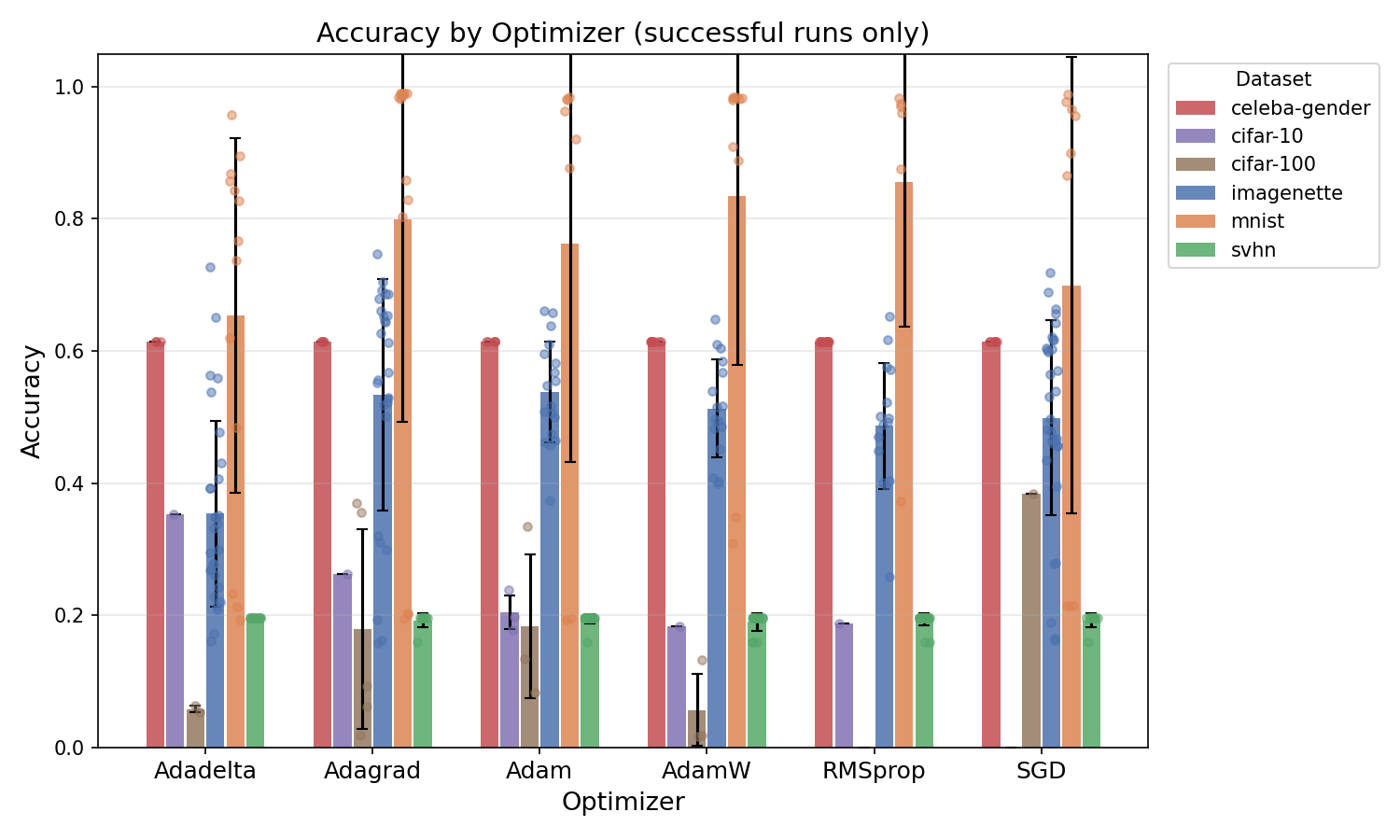}
    \caption{Top-1 accuracy by optimizer. Same variants as
      Fig.~\ref{fig:loss_acc}, grouped by optimizer.}
    \label{fig:opt_acc}
  \end{minipage}
\end{figure*}

\begin{figure}[h]
  \centering
  \includegraphics[width=\linewidth]{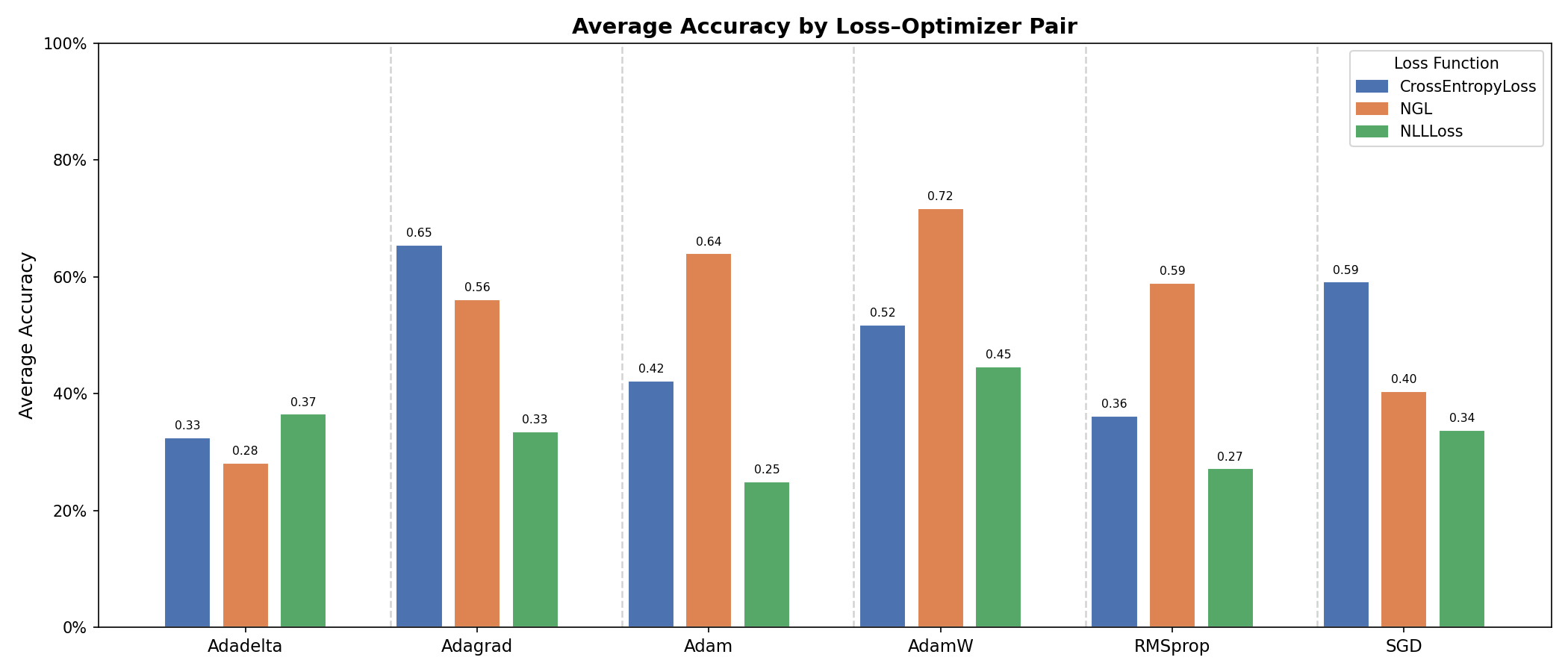}
  \caption{Top-1 accuracy for all 18 loss--optimizer combinations.
    Each cell shows the average accuracy across all successfully trained
    variants for that pairing. Interaction effects are visible as non-uniform patterns across rows, where a loss
    performs differently depending on the optimizer.}
  \label{fig:loss_opt_acc}
\end{figure}

The experimental protocol described in Section~\ref{sec:Experiments}
was executed across the full LEMUR architecture pool.
Before reporting accuracy results, we first characterise how reliably
each pairing trains to completion, since a pairing that silently fails
on a large fraction of architectures cannot be considered robust.
Figure~\ref{fig:loss_opt_acc} shows the average top-1 accuracy for all
18 loss--optimizer combinations, and Table~\ref{table:success} gives
the per-dataset training success rates that contextualise those numbers.

\begin{table}[h]
  \centering
  \fontsize{7.5}{9}\selectfont
  \caption{Per-dataset training success rates.}
  \label{table:success}
  \begin{tabular}{l r}
    \toprule
    \textbf{Dataset} & \textbf{Success rate} \\
    \midrule
    CelebA-Gender & 100.0\% \\
    CIFAR-10      & 100.0\% \\
    CIFAR-100     & 100.0\% \\
    ImageNette    &  22.9\% \\
    MNIST         &  10.3\% \\
    SVHN          &  17.2\% \\
    \bottomrule
  \end{tabular}
\end{table}

\subsection{Training Success}

As shown in Table~\ref{table:success}, CelebA-Gender, CIFAR-10, and
CIFAR-100 achieve a 100\% success rate, confirming that the compatible
architecture subset trains reliably across all 18 pairings on
standard-resolution datasets.
ImageNette, MNIST, and SVHN exhibit substantially lower success rates
(22.9\%, 10.3\%, and 17.2\% respectively), with the remaining failures
dominated by unfixable architecture--dataset mismatches identified in
Section~\ref{sec:compat}.

\subsection{NGL disussion}

NGL achieves competitive accuracy on standard convolutional classifiers
but exhibits its sensitivity to optimizer choice.
If the pipeline utilizes a non-adaptive optimizer such as SGD across all candidate architectures,
NGL becomes an unreliable training recipe and should not be substituted for cross-entropy. 

On the other hand, its strongest results appear with adaptive optimizers Adam and AdamW, 
making these parings the safest choices for NGL.

This makes NGL a conditionally useful tool: superior to NLL and competitive with CEL when paired
with adaptive optimizers, but destructive to training stability under accumulation-based or non-adaptive methods. 

\subsection{Loss Function Comparison}

Figures~\ref{fig:loss_acc}, \ref{fig:opt_acc}, and
\ref{fig:loss_opt_acc} together show that the three loss functions
differ not only in their marginal accuracy distributions but also in
how strongly their performance depends on the choice of optimizer.
Cross-Entropy Loss achieves the highest average accuracy and the lowest
variance across architecture families, consistent with its status as
the standard classification loss.
Its strongest pairings are with Adam and AdamW, while SGD+Momentum
remains competitive but less consistent across the full architecture pool.

NLLLoss performs similarly to CEL on architectures that include an
explicit log-softmax layer but shows higher variance on architectures
where the log-softmax is absent, as the loss receives raw logits
and the gradient signal becomes less stable.
In terms of optimizer sensitivity, NLLLoss follows the same general
pattern as CEL: it is most reliable with Adam and AdamW, remains
serviceable with RMSprop, and is less robust with SGD+Momentum,
Adagrad, and Adadelta.

\subsection{Optimizer Comparison}

Among the six optimizers, Adam and AdamW produce the most consistent
results across all three loss functions and across the evaluated
architecture families.
Both optimizers pair well with CEL and NLLLoss and also provide the
strongest setting for NGL, whose performance otherwise varies more
substantially across optimizers.
Adam and AdamW are the safest default for heterogeneous pools when hyperparameters are fixed.

RMSprop performs comparably to Adam on many standard classifiers, but
its behaviour is less uniform across losses and architectures.
It remains reasonably competitive when paired with CEL and NLLLoss,
but does not stabilise NGL as effectively as Adam or AdamW, and shows
increased instability on deeper architectures with batch normalisation.
RMSprop is therefore better characterised as a mid-tier optimizer:
stronger than the weakest baselines, but not among the most robust
choices across the full loss--optimizer grid.

SGD+Momentum achieves competitive accuracy when paired with
Cross-Entropy Loss on architectures with stable gradient flows on standard classifiers,
but is clearly less reliable for NLLLoss and especially for NGL,
supporting the theoretical prediction by Choi et al.~\cite{Choi2020optimizers}
that less general optimizers may fail where more general ones succeed.

Adagrad and Adadelta consistently underperform the adaptive methods
across all three loss functions, primarily due to the fixed learning
rate (LR~$= 0.01$) in the experimental protocol: Adagrad's
accumulation of squared gradients depletes the effective learning rate
rapidly under this setting, and Adadelta's adaptive scaling converges
slowly without per-model tuning.

\subsection{Best-performing pairings}

The joint accuracy matrix in Figure~\ref{fig:loss_opt_acc} reveals that
the top-performing configurations cluster around adaptive optimizers
paired with CEL or NGL.

Cross-Entropy with Adam and Cross-Entropy with AdamW consistently
achieve the highest average top-1 accuracy across the evaluated datasets,
with AdamW showing a marginal edge on deeper architectures where its
decoupled weight decay provides better regularisation.
NGL paired with Adam and AdamW approaches CEL-level performance on
standard convolutional classifiers such as DenseNet and AirNext,
confirming that NGL is a viable alternative when adaptive optimisation
is available.

SGD+Momentum paired with CEL forms a competitive third tier,
particularly on CIFAR-10 and CIFAR-100 where gradient flows are
well-behaved, but falls behind adaptive methods on ImageNette where
higher resolution and more complex feature distributions complicate
the training process. Such heavy datasets as ImageNette
require much more epochs of training and learning rate scheduling 
for better and faster convergence.

Pairings involving Adagrad or Adadelta rank consistently at the bottom
of the matrix regardless of loss function, and NGL+SGD represents the
single worst-performing combination overall, suggesting that NGL's
non-standard gradient landscape is particularly ill-suited to
non-adaptive optimisation under fixed hyperparameters.

\section{Conclusion}
\label{sec:conclusion}

This paper presented a systematic empirical study of loss--optimizer
pairings across the heterogeneous LEMUR architecture pool, evaluated
under fixed hyperparameters using the automated NNEval framework.
By holding all training settings constant and varying only the
loss--optimizer pair, we isolated the effect of this choice from
confounds introduced by hyperparameter search.

The most novel finding of this work concerns NGL: this study provides
the first systematic evaluation of the genetically evolved loss function.

The results establish clear, actionable preconditions for its use. NGL is a applicable alternative 
to cross-entropy on standard convolutional classifiers, but exclusively when paired with adaptive 
optimizers such as Adam or AdamW. If this condition is not considered, it may cause a significant risk of accuracy degradation, 
particularly on architectures with non-standard gradient flows.

It was found that there is no single pairing is universally optimal
across architecturally diverse models.
Cross-Entropy with Adam or AdamW is the most robust default,
achieving consistently high accuracy across all evaluated datasets and
architecture families.

The injection pipeline scales to any extension of the LEMUR pool 
without per-model manual intervention, providing foundation for future studies.
The most important next step is per-architecture learning rate tuning, particularly 
for Adagrad and Adadelta, whose low rankings here reflect fixed-LR degradation rather 
than inherent weakness.

{
  \small
  \bibliographystyle{ieeenat_fullname}
  \bibliography{bibmain}
}

\end{document}